\documentclass[a4paper,fleqn]{cas-sc}

\usepackage[numbers]{natbib}
\usepackage{graphicx}
\usepackage{amsmath}
\usepackage{amssymb}

\usepackage{amsmath,amsfonts}
\usepackage{algorithmic}
\usepackage{algorithm}
\usepackage{array}
\usepackage{textcomp}
\usepackage{stfloats}
\usepackage{url}
\usepackage{verbatim}
\usepackage{graphicx}
\usepackage{multirow}
\usepackage{amssymb}
\usepackage{amsmath}
\usepackage{amsthm}
\usepackage{booktabs}
\usepackage{arydshln}
\usepackage{float}
\usepackage{color}
\usepackage{colortbl}
\usepackage{xcolor} 
\definecolor{darkgray}{gray}{0.8} 

\def\tsc#1{\csdef{#1}{\textsc{\lowercase{#1}}\xspace}}
\tsc{WGM}
\tsc{QE}

\begin{document}
\let\WriteBookmarks\relax
\def\floatpagepagefraction{1}
\def\textpagefraction{.001}

\shorttitle{Low-Mid Adversarial Perturbation against Unauthorized Face Recognition System}    

\shortauthors{Jiaming Zhang \emph{et al.}}  

\title [mode = title]{Low-Mid Adversarial Perturbation against Unauthorized Face Recognition System}  



%

\author[1]{Jiaming Zhang}[orcid=0000-0003-0991-7109]
\ead{jiamingzhang@bjtu.edu.cn}

\author[1]{Qi Yi}
\ead{21125273@bjtu.edu.cn}

\author[2]{Dongyuan Lu}
\ead{ludy@uibe.edu.cn}
\cormark[1]

\author[1,3]{Jitao Sang}
\ead{jtsang@bjtu.edu.cn}





\affiliation[1]{organization={School of Computer and Information Technology \& Beijing Key Lab of Traffic Data Analysis and Mining, Beijing Jiaotong University},
            city={Beijing},
            postcode={100091}, 
            country={China}}

\affiliation[2]{organization={School of Information Technology and Management, University of International Business and Economics},
            city={Beijing},
            postcode={100029}, 
            country={China}}

\affiliation[3]{organization={Peng Cheng Lab},
            city={Shenzhen},
            postcode={518000}, 
            country={China}}







\cortext[1]{Corresponding author: Dongyuan Lu}



\begin{abstract}
In light of the growing concerns regarding the unauthorized use of facial recognition systems and its implications on individual privacy, the exploration of adversarial perturbations as a potential countermeasure has gained traction. However, challenges arise in effectively deploying this approach against unauthorized facial recognition systems due to the effects of JPEG compression on image distribution across the internet, which ultimately diminishes the efficacy of adversarial perturbations. Existing JPEG compression-resistant techniques struggle to strike a balance between resistance, transferability, and attack potency. To address these limitations, we propose a novel solution referred to as \emph{low frequency adversarial perturbation} (LFAP). This method conditions the source model to leverage low-frequency characteristics through adversarial training. To further enhance the performance, we introduce an improved \emph{low-mid frequency adversarial perturbation} (LMFAP) that incorporates mid-frequency components for an additive benefit. Our study encompasses a range of settings to replicate genuine application scenarios, including cross backbones, supervisory heads, training datasets, and testing datasets. Moreover, we evaluated our approaches on a commercial black-box API, \texttt{Face++}. The empirical results validate the cutting-edge performance achieved by our proposed solutions.
\end{abstract}



\begin{keywords}
 Privacy-preserving \sep Adversarial Attack \sep Face Recognition \sep JPEG
\end{keywords}

\maketitle


\section{Introduction}\label{sec1}

In contemporary times, facial recognition systems (FRS) have witnessed widespread adoption across various applications, such as security access and facial-payment systems~\cite{vo2018robust}. Nevertheless, unauthorized utilization of FRS has consistently raised privacy concerns~\cite{diez2020towards}. For example, Clearview AI, a well-known company, has been revealed to engage in web mining of facial images from social platforms like Facebook and YouTube, and subsequently offer API services to access individuals' facial data~\cite{zhang2020adversarial}. A plethora of scholarly endeavors have focused on protecting users' facial privacy~\cite{yang2023invertible, zhang2020adversarial, cherepanova2020lowkey}, with several proposing the use of adversarial example techniques to shield online-shared facial images from unauthorized FRS, as these examples remain imperceptible to humans but effectively resist unauthorized FRS.

Regrettably, the assumptions underpinning the aforementioned studies tend to be excessively idealistic, overlooking several practical considerations. For instance, when images are uploaded to the internet, they typically undergo image compression, with the predominant method being JPEG compression~\cite{li2022reversible}. This defense mechanism eliminates high-frequency components, consequently weakening the adversarial perturbation. To counteract JPEG compression, several measures have been proposed: (1) some research relies on the principle of Back-Propagation Through Differentiable Approximations (BPDA)~\cite{athalye2018obfuscated}, which employs differentiable mathematical functions to supplant non-differentiable JPEG, or alternatively, trains a generative network to simulate JPEG~\cite{shin2017jpeg}; (2) in contrast, other scholarly pursuits aim to confine the adversarial perturbation exclusively to the image's low-frequency band, primarily based on the assumption that JPEG suppresses noise in the high-frequency spectrum~\cite{zhou2018transferable,SharmaDB19}. Importantly, low-frequency adversarial perturbation is considered more practical as it does not require meticulous attention to specific parameters of JPEG, as in the former case, and has exhibited enhanced transferability across backbones.

Notably, such a solution ceases to be optimal in cases where the source model\footnote{The source model refers to the model that generates the adversarial examples, \emph{i.e.,} surrogate model. The target model refers to the model used to test the adversarial examples.} relies not solely on the low frequency band, but rather employs a range of frequency bands. It has come to light that restricting adversarial perturbation solely to the low-frequency band, while augmenting the ability to withstand JPEG compression, may simultaneously compromise attack performance in broader contexts. Further discussion on this topic can be found in Section~\ref{sec5-analyses}. Therefore, a more sophisticated and efficacious method involves redesigning the source model to integrate low-frequency perturbation, rather than enforcing the adversarial perturbation to reside within a particular frequency band.

In pursuit of this objective, we introduce an innovative \emph{low frequency adversarial perturbations} (LFAP) method via a robust training process. This approach facilitates the source model to primarily harness information within the low-frequency domain, with the aim of generating an adversarial perturbation containing a high concentration of low-frequency components. Moreover, our analysis reveals that mid-frequency components still contribute significantly. Consequently, we combine two models employing relatively low and mid-frequency components to produce \emph{low-mid frequency adversarial perturbations} (LMFAP).

Given that our methodology emphasizes the source model, it can seamlessly integrate with various attack algorithms, such as gradient-based strategies involving input transformation employed by DI~\cite{xie2019improving} and mix-up operations employed by ADMIX~\cite{wang2021admix}. Our assertions have been corroborated through the evaluation of our methods in conjunction with seven pre-existing black-box attack algorithms.

To compellingly demonstrate the effectiveness of our approach in real-world situations, we designed an array of black-box configurations encompassing cross-backbones, supervisory heads, training datasets, and testing datasets of FRS. Additionally, we assessed our method on a commercial black-box API, specifically \texttt{Face++}\footnote{\url{https://www.faceplusplus.com/}}. To the best of our knowledge, it is worth noting that our research represents the first attempt where adequate evaluation tools have been utilized. 

The contributions of this paper are multifaceted.

\begin{itemize}
\item We put forward a novel technique to generate low-mid adversarial perturbations that are resilient to JPEG compression. These perturbations are aimed at safeguarding the privacy of face images uploaded to social media platforms from the prying eyes of unauthorized FRS.

\item We introduced the LFAP approach through a robust training process and devised ensemble adversarial attacks to propose the refined LMFAP.

\item We demonstrated that our methods (both LFAP and LMFAP) can be seamlessly integrated with existing black-box attack algorithms, significantly enhancing their performance across a diverse array of black-box configurations and the commercial API.

\end{itemize}

\section{Background and Related Work}\label{sec2}

\subsection{Preliminaries}
A complete FRS consists of two modules: the backbone network $f$ used to extract features, such as ResNet50~\cite{he2016deep}, and the supervisory head $h$ used during training, such as ArcFace~\cite{deng2019arcface}. Given an original image $\mathbf{x}$, $f$ outputs an $l$-dimension embedding $f(\mathbf{x})$. In the training phase, the FRS is optimized by an objective function $\mathcal{J}(h(f(\mathbf{x})), y)$, and $y$ is the identity corresponding to $\mathbf{x}$. In the testing phase, $\mathbf{x_e}$ represents the enrolled image corresponding to $\mathbf{x}$, \emph{i.e.}, $\mathbf{x_e}$ and $\mathbf{x}$ are a pair of images belonging to the same subject. When the distance $d(f(\mathbf{x}), f(\mathbf{x_e}))$ between the two images is less than a certain threshold, they are judged to be the same subject, otherwise they are different subjects. Note that the fully-connected layer in the supervisory head becomes redundant once the training is completed. As classification models are bound by the number of distinct categories, there is little discussion of the traditional black-box adversarial attacks on crossing the training set, \emph{i.e.}, the difference between the training set of the source model and the target model. However, FRS is not restrained by the number of categories and hence, crossing the training set often becomes a more common scenario in real-world applications. Nevertheless, previous works have overlooked this aspect and have not introduced the concept of different training datasets in the black-box experiments.

\subsection{Adversarial Attack}
Adversarial attacks (or examples) can fool deep neural networks with very small image noise~\cite{szegedy2013intriguing}. To obtain the adversarial perturbations $\delta$, we consider $\ell_\infty$-norm constrained perturbation in this work, where $\delta$ satisfies $\Vert \delta \Vert_\infty \leq \epsilon$ with $\epsilon$ being the maximum perturbation magnitude. The adversarial black-box attacks can be categorized into two categories: (1) query-based attack needs the feedback of iterative queries to target models, which is not consistent with the assumptions of this paper; (2) transferability-based attack that use the adversarial examples generated on some surrogate models to attack the other target models~\cite{lin2023sensitive}. The most common of this type of attacks are gradient-based methods such as FGSM~\cite{goodfellow2014explaining}, MI~\cite{dong2018boosting}, DI~\cite{xie2019improving}, TI~\cite{dong2019evading}, SI~\cite{lin2019nesterov}, ADMIX~\cite{wang2021admix} and some combinations of them. We provide a brief overview of FGSM to illustrate this series of techniques:
\begin{equation}\label{eq1}
  \delta = \epsilon \cdot {\rm{sign}}(\nabla_{\mathbf{x}} \mathcal{L}(\mathbf{x}, \mathbf{x_{e}}; \theta)),
\end{equation}
where $\theta$ is the parameter of network, $\mathcal{L}$ is the loss function at the testing phase, usually a negative distance function, and the Euclidean distance is employed as the distance function in this work.

Liu \emph{et al.}~\cite{liu2016delving} were the first to suggest incorporating ensemble learning in adversarial attacks. Literature suggests a unanimous agreement that the greater the difference in the structure of the ensemble models, the better the generated adversarial examples become~\cite{zhang2021robust}. It is noteworthy that our ensemble adversarial attacks employ an identical architecture, in contrast to the conventional ensemble adversarial attacks. This brings to light an alternative method to harness ensemble learning in adversarial attacks. Furthermore, there are some model-specific designed attacks, such as LinBP~\cite{guo2020backpropagating}, which modifies the activation neurons during the back-propagation, and TTP~\cite{naseer2021generating}, which has a specific generative model. Nonetheless, it is not a straightforward task to extend these findings to other architectures or modules.

\subsection{Resisting JPEG Compression in Face Recognition System}
JPEG compression is the most common image compression technique used for image transmission over the web, and is also an adversarial defense~\cite{naseer2021generating}. Even though Zhang \emph{et al.}~\cite{zhang2020adversarial} first proposed the use of adversarial example to protect face privacy as a solution, both APF~\cite{zhang2020adversarial} and LowKey~\cite{cherepanova2020lowkey} only consider improving attack performance, while disregarding the impact of JPEG compression. In this regard, Athalye \emph{et al.}~\cite{athalye2018obfuscated} pointed out that attackers can approximate the non-differentiable JPEG with a differentiable module and regard it as a sub-module of the entire model, and thus derive the gradient-based adversarial perturbation. Shin \emph{et al.}~\cite{shin2017jpeg} replaced non-differentiable JPEG with differentiable mathematical functions. Wang \emph{et al.}~\cite{wang2020towards} trained a generative network to simulate JPEG. However, this class of methods can only target JPEG compression of specific factors, \emph{e.g.}, an adversarial example generated for a quality factor of $50$ will fail when faced with a JPEG compression with a quality factor of $75$. In contrast, other works tried to restrict the perturbation to the low frequency band, which avoids the effect of JPEG compression. Zhou \emph{et al.}~\cite{zhou2018transferable} used a form of low-pass filter to enforce the adversarial perturbation into low frequency band. Yash \emph{et al.}~\cite{SharmaDB19} removed certain high frequency components of the perturbation by the discrete cosine transform (DCT) and inverse discrete cosine transform (IDCT). However, these methods weaken the performance of adversarial perturbation in some cases.

\begin{figure}[t]
\begin{minipage}[b]{0.32\linewidth}
\centering
\includegraphics[width=0.99\textwidth]{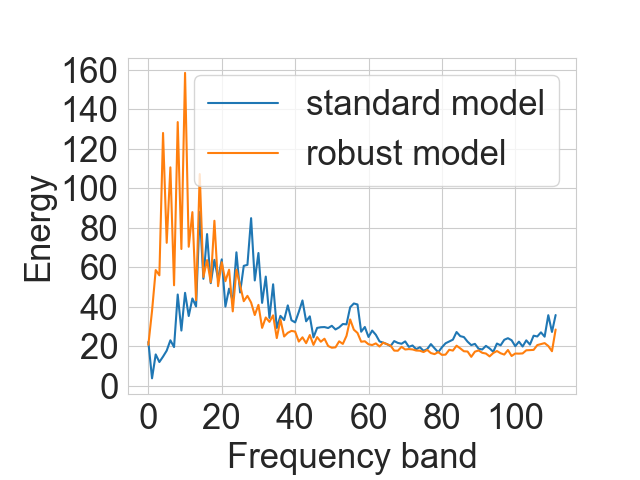}
\centerline{(a)}
\end{minipage}
\begin{minipage}[b]{0.32\linewidth}
\centering
\includegraphics[width=0.99\textwidth]{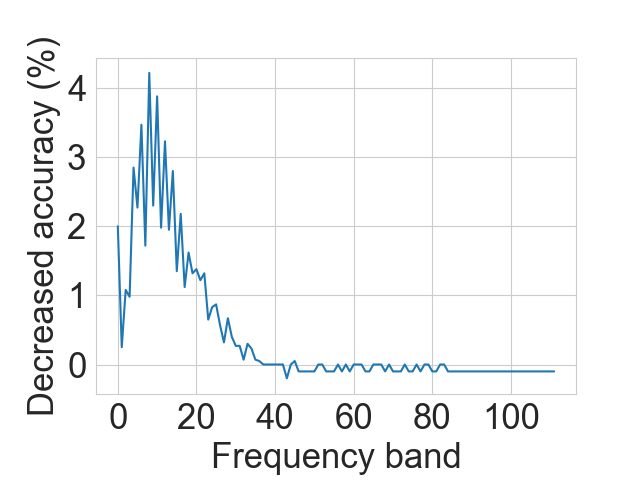}
\centerline{(b)}
\end{minipage}
\begin{minipage}[b]{0.32\linewidth}
\centering
\includegraphics[width=0.99\textwidth]{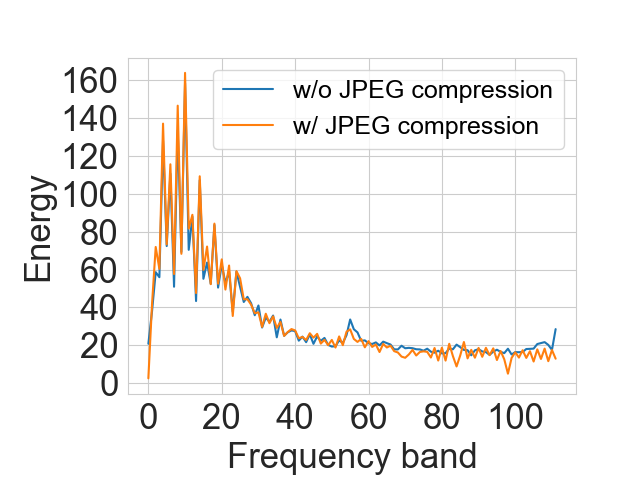}
\centerline{(c)}
\end{minipage}
\caption{(a) The robust model and standard model utilize different frequency components. (b) Removing different frequency components leads to different decreased accuracy. (c) JPEG compression weakens mainly the high frequency band.}
 \label{fig3}
\end{figure}

\begin{figure*}[t]
\begin{minipage}[b]{0.49\linewidth}
\centering
\includegraphics[width=0.99\textwidth]{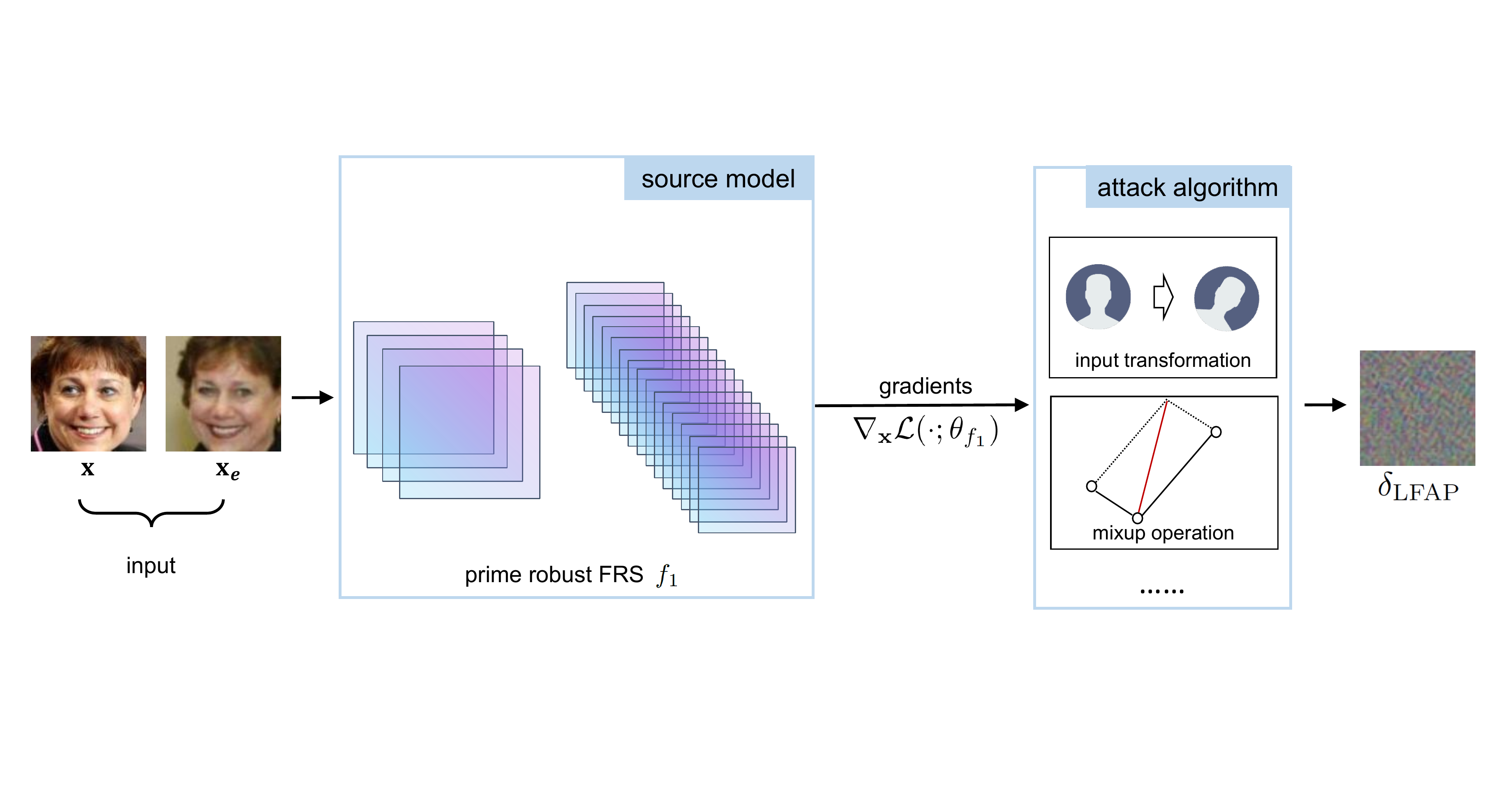}
\centerline{(a) LFAP}
\end{minipage}
\begin{minipage}[b]{0.49\linewidth}
\centering
\includegraphics[width=0.99\textwidth]{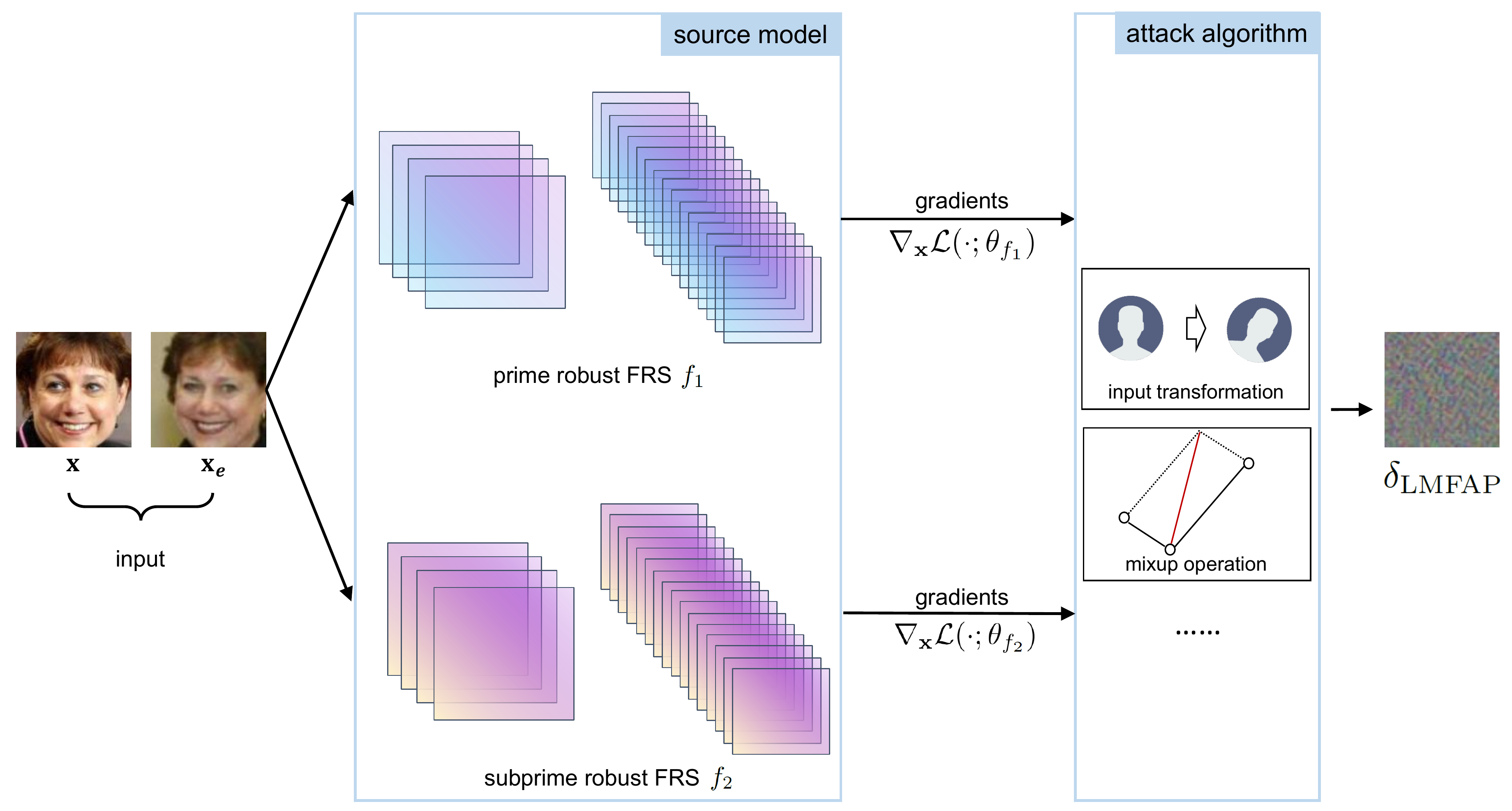}
\centerline{(b) LMFAP}
\end{minipage}
\caption{The pipeline framework. $\mathbf{x}$ is the face image to be protected and $\mathbf{x_e}$ is the enrolled image of the same subject as $\mathbf{x}$. The generated perturbation $\delta_{\rm{LFAP}}$ and noise $\delta_{\rm{LFMAP}}$, when added to $\mathbf{x}$, resist JPEG compression and prevent the unauthorized face recognition model from matching the adversarial image $\mathbf{x}+\delta$ with the enrolled image $\mathbf{x_e}$.}
 \label{fig-framework}
\end{figure*}

\section{Low Frequency Adversarial Perturbation (LFAP)}\label{sec4}

\subsection{Robust Model Utilizing Low Frequency Information}\label{sec4-1}


Prevailing literature substantiates a potent positive association between the behavior of models capitalizing on robust features in images and those contingent upon their low-frequency components~\cite{wang2020frequency,wang2020high,chen2021amplitude}. To authenticate the practicability of engendering low-frequency adversarial perturbations derived from a robust source model, we meticulously examined the frequency constituents intrinsic to such modifications. Distinctly, we trained a conventional model and a robust counterpart employing the ImageNet training dataset~\cite{russakovsky2015imagenet}, respectively. The robust model underwent adversarial training within the radius-4 neighborhood. Subsequently, we contrasted the disparities in frequency domain constituents of the adversarial perturbations generated by the two models on the validation set. To determine the frequency domain constituents of the adversarial perturbations, we employed the DCT technique to gauge the energy variation between the original and the adversarial images. The aggregate energy for the identical frequency band is depicted in Figure~\ref{fig3}(a). It is remarkable that the robust model spawns an increased volume of perturbations in the low-frequency spectrum, diverging from the conventional model. This revelation prompted us to utilize a robust FRS to generate the desired adversarial perturbations.


\subsection{Methodology}\label{sec4-2}

Pursuing the concept of employing a robust model to engender low-frequency adversarial perturbations, we conduct conventional adversarial training~\cite{madry2018towards} on the FRS, as depicted in Equation~\eqref{eq2}:
\begin{equation}\label{eq2}
  \min_{\theta_f, \theta_H} \mathbb{E}_{(\mathbf{x}, y)\sim D} [\max_{\Vert \delta \Vert_{\infty \leq \epsilon}} \mathcal{J}(h(f(\mathbf{x}+\delta)), y; \theta_f, \theta_H)],
\end{equation}
where $\theta_f$ and $\theta_h$ represent the parameters of the backbone $f$ and the head $h$, respectively. This approach enables the model to primarily exploit information within the low-frequency spectrum, anticipating that the generated adversarial perturbation on the robust model encompasses a greater low-frequency constituent. Note that once the training of a robust FRS is completed, the head $h$ can be discarded. Consequently, we have procured the prime robust FRS $f_1$ with $\theta_{f_1}$. Thus, exemplifying with FGSM, the generation procedure for Low-Frequency Adversarial Perturbation (LFAP) solely necessitates the substitution of parameters $\theta$ in Equation~\eqref{eq1} with the parameters $\theta_{f_1}$ acquired in Equation~\eqref{eq2}, as demonstrated in Equation~\eqref{eq3}:
\begin{equation}\label{eq3}
  \delta_{\rm{LFAP}} = \epsilon \cdot {\rm{sign}}(\nabla_{\mathbf{x}} \mathcal{L}(\cdot; \theta_{f_1})).
\end{equation}
A visual representation of this generation is depicted in Figure~\ref{fig-framework}(a). After the training of $f_1$ is performed, the gradient $\nabla_{\mathbf{x}} \mathcal{L}$ with respect to the input ($\mathbf{x}$ and $\mathbf{x_e}$) and $f_1$ is fed to the alternative attack algorithm, resulting in the acquisition of $\delta_{\rm{LFAP}}$. For instance, if the attack algorithm is a DI-MI that uses the input transformation, then the LFAP can be denoted as DI-MI-LFAP.

\section{Low-Mid Frequency Adversarial Perturbation (LMFAP)}\label{sec5}

\subsection{Contribution of Mid Frequency Components}\label{sec5-1}

In an endeavor to determine the efficacy of low-frequency perturbations as the optimal solution, we conducted an exhaustive analysis of the impact of full-frequency domain components on FRS. This was achieved by transforming the image $\mathbf{x}$ into the frequency domain via the Discrete Cosine Transform (DCT) and subsequently applying a frequency mask to eliminate specific frequency domain components. The image $\mathbf{x}_{\rm{masked}}$ was then reconstructed from the spectrum matrix ${\rm{DCT}}(\mathbf{x})$ utilizing the Inverse Discrete Cosine Transform (IDCT) as illustrated in Equation~\eqref{eq4}:
\begin{equation}\label{eq4}
  \mathbf{x}_{\rm{masked}} = {\rm{IDCT}}({\rm{DCT}}(\mathbf{x})\cdot M),
\end{equation}
where $M$ is the mask matrix to remove $n$-th component, which can be expressed in the following form:
\begin{equation}\label{eq5}
  M_{i,j}=
\begin{cases}
0,& i=n \; \text{or} \; j=n\\
1,& \text{else}
\end{cases}.
\end{equation}

By examining the diminished accuracy of the reconstructed images $\mathbf{x}_{\rm{masked}}$ on the FRS, we ascertained the contribution of particular frequency components to the system. Consequently, we assessed the reduced accuracy correlated to $n \in [1, 112]$, depicted in Figure~\ref{fig3}(b). Our investigation revealed that the low-frequency band component (approximately $[1, 20]$ in the LFW dataset~\cite{huang2008labeled}) in the image substantially contributed to the model, thereby offering substantial experimental support for the implementation of low-frequency adversarial perturbations, as delineated in the previous subsection. However, we also observed a noteworthy contribution from the mid-frequency band (approximately $[20, 40]$ in the LFW dataset) component. As illustrated in Figure~\ref{fig3}(a), the robust model exhibited a lack of proficiency in utilizing the mid-frequency component. This observation prompted us to amalgamate these two components to augment the performance of adversarial perturbation. Moreover, we evaluated the influence of JPEG compression on adversarial perturbation by comparing adversarial perturbations with and without JPEG compression, as demonstrated in Figure~\ref{fig3}(c). Our findings indicated that JPEG compression predominantly attenuated the high-frequency band ($>50$), thus substantiating the feasibility of employing low-mid frequency adversarial perturbations.

\subsection{Methodology}\label{sec5-2}

In the preceding section, we explicated that the designated model accounts for the mid-frequency component present within the image, which remains predominantly resilient to JPEG distortion. Consequently, it might be tenable to devise an adversarial perturbation incorporating a segment of the mid-frequency component in conjunction with the low-frequency component to augment the resulting output. The critical aspect resides in effectively amalgamating both the low and mid-frequency components into the perturbation.

Drawing inspiration from ensemble learning in adversarial attacks~\cite{liu2016delving}, we require one model that uses low-frequency information and another one that leverages mid-frequency information. So we first obtain the prime robust FRS $f_1$ with $\theta_{f_1}$ by adversarially training $m_1$ epochs in the radius-$\epsilon_1$ neighborhood via Equation~\eqref{eq2}. Second, still via Equation~\eqref{eq2}, but we adversarially train $m_2$ epochs in the radius-$\epsilon_2$ neighborhood to obtain the subprime robust FRS $f_2$ with $\theta_{f_2}$. The values of $\epsilon_2$ and $m_2$ corresponding to $f_2$ should be smaller compared to the $\epsilon_1$ and $m_1$ in order to obtain the model that utilizes relatively mid frequency information. Instead of simply summing the adversarial perturbation generated by the $f_1$ and $f_2$, we use ensemble learning to integrate them. Using FGSM as the example, the generation process of \emph{Low-Mid Frequency Adversarial Perturbation} (LMFAP) can be characterized as follows:
\begin{equation}\label{eq6}
  \delta_{\rm{LMFAP}} = \epsilon \cdot {\rm{sign}}(\nabla_{\mathbf{x}} [\mathcal{L}(\mathbf{x}, \mathbf{x_{e}}; \theta_{f_1}) + \lambda \cdot \mathcal{L}(\mathbf{x}, \mathbf{x_{e}}; \theta_{f_2})]),
\end{equation}
where $\lambda$ is a hyper-parameter that balances the contributions of the two FRS. The illustration of the generation is shown in Figure~\ref{fig-framework}(b). After the training of $f_1$ and $f_2$ is performed respectively, the gradients $\nabla_{\mathbf{x}} \mathcal{L}(\mathbf{x}, \mathbf{x_{e}}; \theta_{f_1})$ and $\nabla_{\mathbf{x}} \mathcal{L}(\mathbf{x}, \mathbf{x_{e}}; \theta_{f_2})$ are fed to the alternative attack algorithm to obtain $\delta_{\rm{LMFAP}}$.

\begin{figure}[t]
  \centering
  \includegraphics[width=0.7\linewidth]{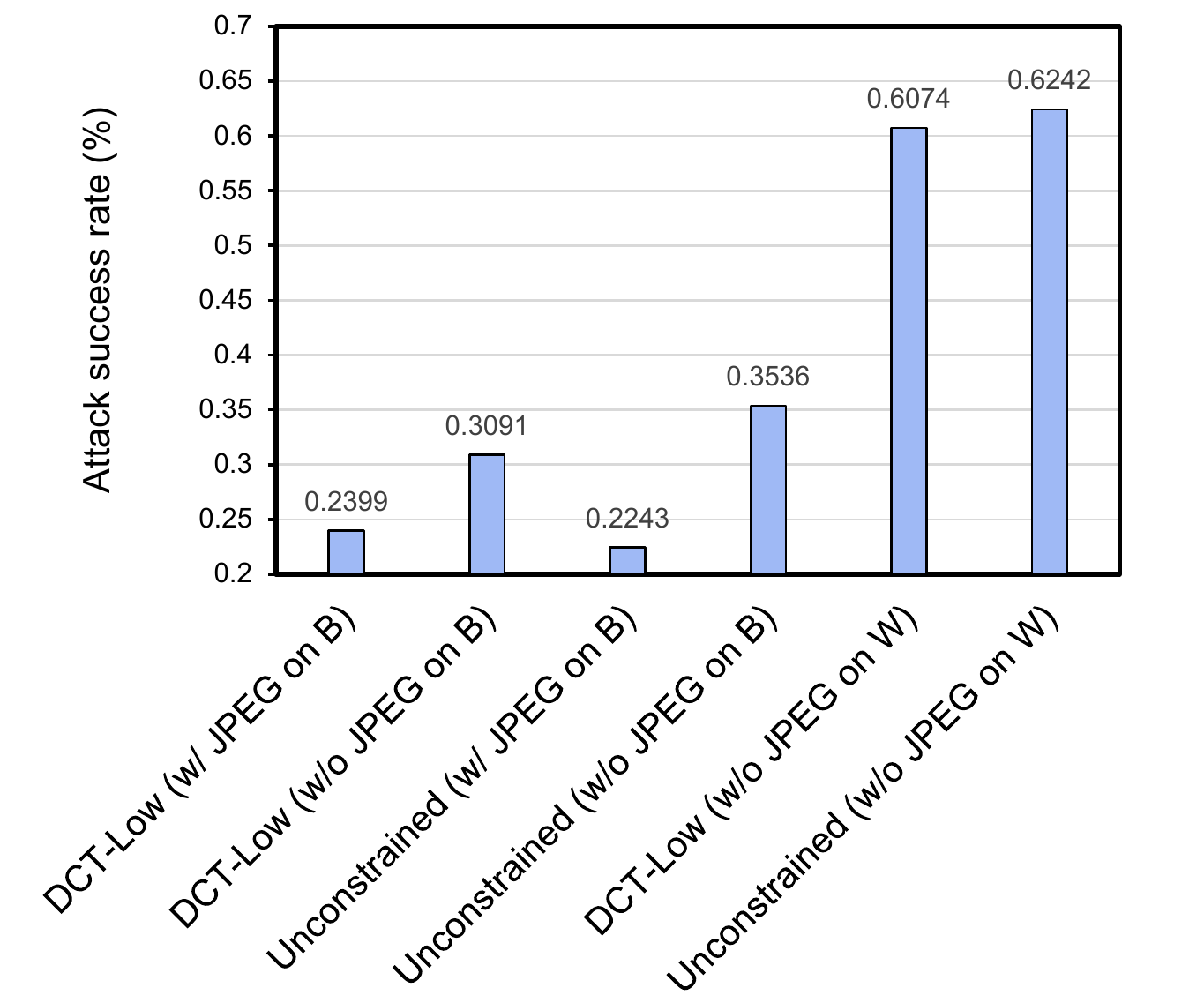}
  \caption{(b) The success rates of black-box attacks (B) and white-box attacks (W) are associated with different forms of adversarial perturbations. DCT-Low is an adversarial perturbation that has had high-frequency components removed, while Unconstrained denotes the original adversarial perturbation. The JPEG compression quality factor is set to 50.}\label{fig2-a}
\end{figure}

\begin{figure}[t]
  \centering
  \includegraphics[width=0.8\linewidth]{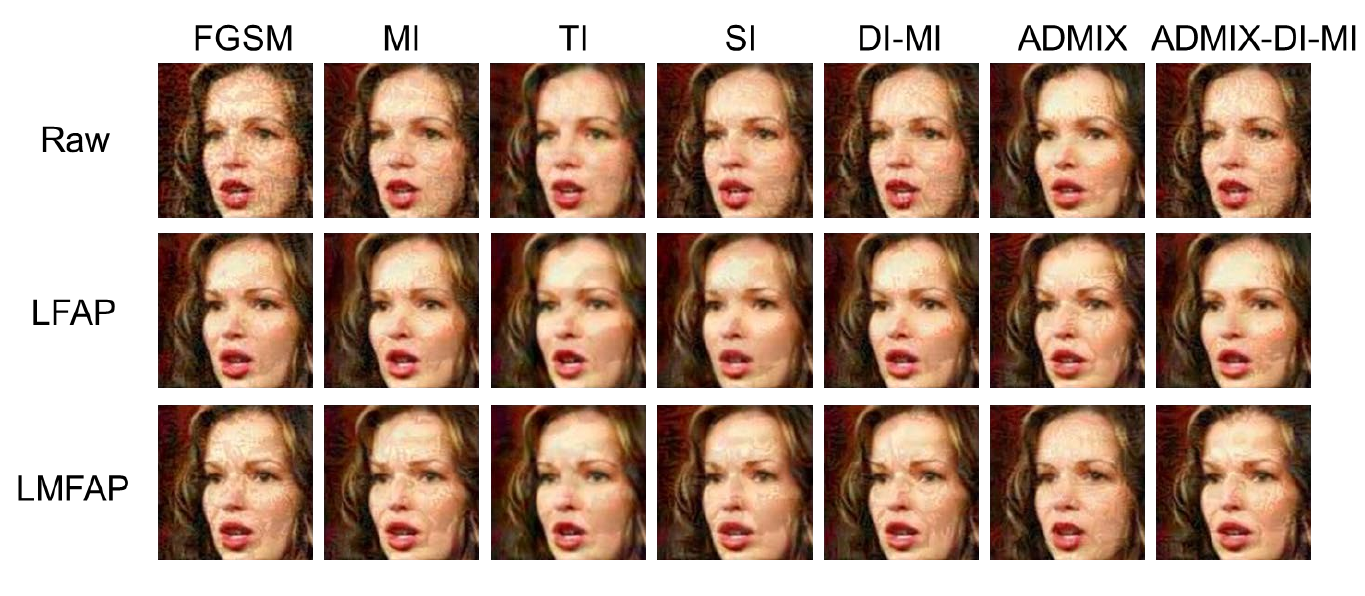}
  \caption{The provided adversarial images are demonstrated in seven columns that correspond to seven distinct baseline attacks. The initial row relates to the raw baseline attack, followed by the attacks involving LFAP and LMFAP, respectively, in the second and third rows.}\label{fig2-b}
\end{figure}

\begin{figure}[t]
  \centering
  \includegraphics[width=0.9\linewidth]{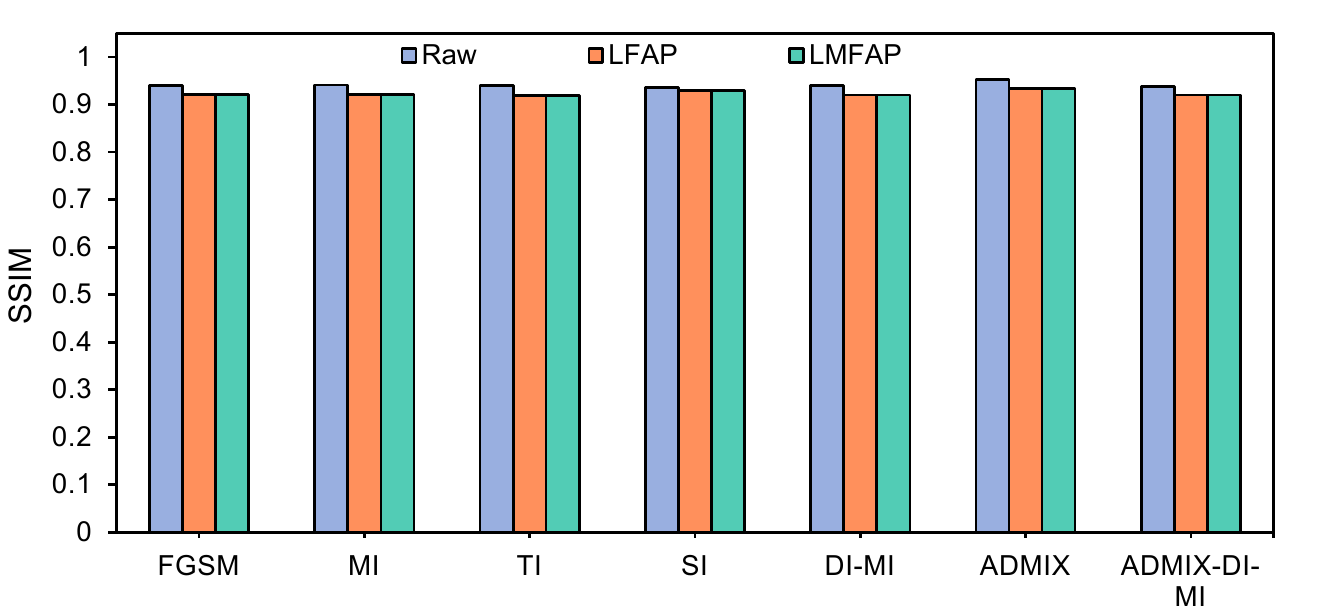}
  \caption{The SSIM value of LFW dataset between the generated adversarial images and original images. A higher value of the SSIM indicates a higher quality of the adversarial images that are generated.}\label{fig_ssim}
\end{figure}

\section{Experiments}

\subsection{Setups.}

To extensively evaluate the practical performance of our methods, with the help of the {\tt FaceX-Zoo} toolbox\footnote{\url{https://github.com/JDAI-CV/FaceX-Zoo/}}, we prepared several models with different backbones, supervisory heads, training datasets as target models. 

\paragraph{Datasets.} 
We train the source FRS on CASIA-WebFace dataset~\cite{yi2014learning}. The FRS used to test the adversarial examples (\emph{i.e.,} target models) are trained on one of three datasets: MS-Celeb-1M (MS)~\cite{guo2016ms}, CASIA-WebFace (CA) and VGGFace2 (VG)~\cite{cao2018vggface2}. LFW~\cite{huang2008labeled}, AgeDB-30~\cite{moschoglou2017agedb} and CFP-FP~\cite{sengupta2016frontal} datasets are used as test datasets to verify the effectiveness of the adversarial examples. MS-Celeb-1M contains 10M images of 100K subjects for training. CASIA-WebFace contains 494,414 images of 10,575 subjects for training. With the same setup as \cite{zhang2020adversarial}, we aim to jam the unauthorized FRS to match the same users, thus the positive pairs (belong to the same person) are used for testing. In the section, we select subjects with two face images, where one is used as the enrolled image, and the other is the original image for synthesizing the adversarial image. LFW contains 13,233 images of 5,749 different subjects. According to the refined version of \cite{deng2019arcface}, we use 6,000 images to construct 3,000 positive pairs of images. AgeDB-30 contains 16,488 images of 568 different subjects. Same as above, we use 6,000 images to construct 3,000 positive pairs of images. CFP-FP contains 7,000 images of 500 different subjects. Same as above, we use 7,000 images to construct 3,500 positive pairs of images.

\paragraph{Models.}
For backbones, we consider TF-NAS-A~\cite{hu2020tf}, GhostNet~\cite{han2020ghostnet}, MobileFaceNet (MFNet)~\cite{chen2018mobilefacenets}, Inception-ResNet-v1 (In-Res-v1)~\cite{szegedy2017inception}, ResNet50-IR-SE (Res50-I-S)~\cite{he2016deep}, ResNet50-IR (Res50-I)~\cite{hu2018squeeze}, SwinTransformer (SwinT)~\cite{liu2021swin} and Sphere20a (Sphere)~\cite{coors2018spherenet}. For heads, we consider ArcFace (Arc)~\cite{deng2019arcface}, FaceNet (FN)~\cite{schroff2015facenet}, SphereFace (Sphere)~\cite{liu2017sphereface} and MV-Softmax (MV)~\cite{wang2020mis}. Therefore, a FRS can be denoted as {\sc backbone@head@training dataset}. \emph{E.g.,} the FRS trained on CASIA-WebFace dataset and employing ResNet50-IR-SE backbone and MV-Softmax head can be denoted as ResNet50-I@MV@CA.

\paragraph{Baselines.}
To show the performance improvement of our methods incorporated with black-box attack algorithms, we employ FGSM, MI, TI, SI, DI-MI, ADMIX and ADMIX-DI-MI as \emph{black-box attack} baselines. To demonstrate the superiority of our methods to resist JPEG compression, we consider Differentiable-JPEG~\cite{shin2017jpeg}, ComReAdv~\cite{wang2020towards}, TAP~\cite{zhou2018transferable}, DCT-Low~\cite{SharmaDB19}, LowKey~\cite{cherepanova2020lowkey} and APF~\cite{zhang2020adversarial} as \emph{resisting JPEG attack} baselines.

\paragraph{Hyper-parameters.}
For prime robust FRS $f_1$, we adopt radius-$4$ neighborhood adversarial training by SGD optimizer with $0.9$ momentum for $50$ epochs with the initial learning rate of $0.1$ divided by $10$ at Epoch $30$ and $45$, respectively. For subprime robust FRS $f_2$, we adopt radius-$1$ neighborhood adversarial training by SGD optimizer with $0.9$ momentum for $20$ epochs with the initial learning rate of $0.1$ divided by $10$ at Epoch $10$, respectively, \emph{i.e.}, $\epsilon_1=4, \epsilon_2=1, m_1=50, m_2=20$. The maximum perturbation $\epsilon$ of each pixel is set to $16$ unless otherwise stated, the number of iterations is set to $10$, and the step size is set to $1.25$. The $\lambda$ in Equation~\eqref{eq6} is set to $0.6$. We discuss the hyper-parameter in Section~\ref{sec5-5}.

\paragraph{Evaluation Metrics.}
\begin{itemize}
    \item \textbf{Attack success rate (ASR)} is used to measures the effectiveness of adversarial images in this work: 
    
    \begin{equation}\label{eq11}
        {\rm{ASR}} = \frac{N_{w/o} - N_{w/}}{N_{total}}
    \end{equation}
    where $N_{w/o}$ and $N_{w/}$ denote the number of correctly recognized original images and adversarial images, respectively. $N_{total}$ denotes the total number of images. The higher ASR value indicates the effectiveness of the adversarial image.
    
\end{itemize}    

\begin{itemize}
    
    \item \textbf{SSIM} is used as a normalized metric in this work, which quantifies the similarity between the perturbed images and the original images~\cite{wang2004image}:
    
        \begin{equation}\label{eq13}
        {\rm{SSIM}}(x,y)=\frac{(2\mu_x\mu_y+c_1)(2\sigma_{xy}+c_2)}{(\mu_x^2+\mu_y^2+c_1)(\sigma_x^2+\sigma_y^2+c_2)}
    \end{equation}
    
    According to \cite{wang2003multiscale}, $x$ and $y$ are the two images to be compared, $\mu_x,\, \mu_y,\, $ $\sigma_x, \sigma_y,\, \sigma_{xy}$ are the means and variances of $x$ and $y$, the covariance of $x$ and $y$, respectively. $c_i = (d_iJ)^2$, where $J=(2^{({\rm \sharp \ of \ bits \ per \ pixel})}-1), \,  d_1=0.01, \,d_2=0.03$.
    
\end{itemize}

\begin{table*}[t]
\centering
\caption{The success rates of black-box attacks (\%) across various datasets and under varying levels of JPEG compression. The values with a gray background were obtained using the proposed methods. The top-1 best results are highlighted in \textbf{bold}.}
\label{tab2}
\resizebox{1.0\linewidth}{!}{
\setlength{\tabcolsep}{1.0mm}{
\begin{tabular}{c|ccc|ccc|ccc} 

 \hline
 & \multicolumn{3}{c|}{LFW} & \multicolumn{3}{c|}{AgeDB-30} & \multicolumn{3}{c}{CFP-FP} \\
  \cline{2-10} 
  Attacks & W/O & Quality=75 & Quality=50 & W/O & Quality=75 & Quality=50 & W/O & Quality=75 & Quality=50 \\
  \hline
  Differentiable-JPEG-75 & 72.30 & 73.87 & 70.10 & 61.94 & 61.40 & 59.54 & 47.31 & 47.48 & 43.60 \\
  Differentiable-JPEG-50 & 73.60 & 73.83 & 72.20 & 63.54 & 62.27 & 61.67 & 47.40 & 47.08 & 48.40 \\
  ComReAdv-75            & 50.04 & 48.13 & 47.57 & 47.64 & 47.77 & 45.34 & 37.54 & 36.65 & 34.94 \\
  ComReAdv-50            & 49.67 & 47.73 & 47.53 & 48.24 & 46.90 & 45.77 & 36.65 & 36.85 & 35.74 \\
  TAP                    & 74.43 & 74.20 & 70.66 & 53.87 & 52.30 & 52.27 & 41.25 & 39.45 & 39.69 \\
  DCT-Low                & 68.00 & 70.70 & 67.57 & 61.90 & 59.30 & 58.64 & 46.05 & 45.05 & 41.63 \\
  LowKey                 & 83.41 & 80.70 & 78.70 & 80.14 & 77.26 & 74.47 & 66.86 & 64.03 & 61.40 \\
  APF                    & 91.90 & 90.42 & 89.61 & 80.11 & 78.34 & 77.24 & 48.95 & 45.72 & 44.23 \\
  \hline
  \rowcolor{darkgray}
  LFAP                   & 77.07 & 75.66 & 75.96 & 67.10 & 66.50 & 65.94 & 47.68 & 48.02 & 46.26 \\
  \rowcolor{darkgray}
  LMFAP                  & \textbf{96.87} & \textbf{96.26} & \textbf{95.86} & \textbf{86.34} & \textbf{85.37} & \textbf{84.87} & \textbf{74.88} & \textbf{75.37} & \textbf{73.49} \\
\hline
\end{tabular}}}
\end{table*}

\begin{table*}[t]
\centering
\caption{The black-box attack success rates (\%) on LFW dataset under $\epsilon=16$. The values with a gray background were obtained using the proposed methods. The top-1 best results are highlighted in \textbf{bold}.}
\label{tab1}
\resizebox{1.0\linewidth}{!}{
\setlength{\tabcolsep}{2.8mm}{
\begin{tabular}{c|ccccccccc} 

 \hline
 \multirow{3}{*}{Attacks} & TF-NAS-A & GhostNet & SwinT & MFNet & MFNet & MFNet & In-Res-v1 & ResNet50-I-S & Sphere \\
 & @MV & @MV & @MV & @MV & @MV & @Arc & @FN & @Arc & @SF \\
 & @MS & @MS & @MS & @MS & @CA & @CA & @VG & @MS & @CA \\
  \hline
  FGSM       & 29.05 & 29.36 & 15.61 & 31.80 & 72.66 &44.38 & 33.30 & 27.68 & 62.05\\
  \rowcolor{darkgray}
  FGSM-LFAP  & 39.51 & 33.96 & 29.84 & 33.59 & 58.33 &48.98 & 48.36 & 35.88 & 47.11\\
  \rowcolor{darkgray}
  FGSM-LMFAP & 64.15 & 59.76 & 49.71 & 59.63 & 83.63 &73.01 & 64.56 & 62.28 & 66.11\\
  \hline
  MI       & 16.21 & 16.30 &  6.61 & 16.33 & 48.26 &27.98 & 15.33 & 15.41 & 32.65\\
  \rowcolor{darkgray}
  MI-LFAP  & 51.08 & 46.16 & 40.35 & 44.26 & 71.09 &62.61 & 55.13 & 47.58 & 56.68\\
  \rowcolor{darkgray}
  MI-LMFAP & 83.41 & 79.16 & 71.55 & 78.83 & 93.46 &89.68 & 77.59 & 82.78 & 81.08\\
\hline
  TI       &  7.28 &  7.16 &  3.15 &  7.56 & 37.36 &12.18 & 17.43 & 7.48 & 19.08\\
  \rowcolor{darkgray}
  TI-LFAP  & 15.15 & 16.16 &  8.11 & 13.30 & 41.20 &21.41 & 42.73 & 14.98 & 27.78\\
  \rowcolor{darkgray}
  TI-LMFAP & 43.61 & 42.30 & 26.91 & 37.86 & 78.23 &53.25 & 67.63 & 44.78 & 53.85\\
  \hline
  SI       & 22.55 & 24.16 & 11.55 & 27.63 & 53.90 & 38.78 & 30.46 & 22.05 & 55.05\\
  \rowcolor{darkgray}
  SI-LFAP  & 48.08 & 43.53 & 36.05 & 41.66 & 67.19 & 60.11 & 52.26 & 44.28 & 54.65\\
  \rowcolor{darkgray}
  SI-LMFAP & 83.18 & 78.93 & 69.81 & 77.90 & 92.79 & 89.41 & 75.36 & 81.71 & 79.78\\
  \hline
  DI-MI       & 22.61 & 22.43 &  9.44 & 22.73 & 57.43 &34.71 & 21.46 & 22.91 & 40.88\\
  \rowcolor{darkgray}
  DI-MI-LFAP  & 53.48 & 48.49 & 41.61 & 45.99 & 72.36 &64.68 & 59.20 & 49.34 & 59.54\\
  \rowcolor{darkgray}
  DI-MI-LMFAP & 86.01 & 82.03 & 75.05 & 81.03 & 94.53 &91.45 & 81.23 & 85.58 & 84.08\\
  \hline
  ADMIX      & 15.08 & 16.76 &  5.68 & 18.43 & 52.16 &29.41 & 14.49 & 16.68 & 31.25\\
  \rowcolor{darkgray}
  ADMIX-LFAP & 53.21 & 47.90 & 39.65 & 45.53 & 73.46 &64.51 & 54.46 & 46.81 & 58.71\\
  \rowcolor{darkgray}
  ADMIX-LMFAP& 88.11 & 85.46 & 77.75 & 84.50 & 95.89 &93.51 & 81.23 & 88.11 & 86.81\\
\hline
  ADMIX-DI-MI      & 33.28 & 32.36 & 16.65 & 34.43 &71.90 & 49.55 & 29.16 & 33.18 & 47.45\\
  \rowcolor{darkgray}
  ADMIX-DI-MI-LFAP & 59.08 & 52.79 & 46.28 & 49.96 &75.76 & 68.08 & 63.16 & 53.68 & 61.84\\
  \rowcolor{darkgray}
  ADMIX-DI-MI-LMFAP& \textbf{89.35} & \textbf{87.33} & \textbf{79.54} & \textbf{86.20} & \textbf{96.19} &\textbf{94.08} & \textbf{84.96} & \textbf{90.31} & \textbf{88.01}\\
\hline
\end{tabular}}}
\end{table*}

\begin{table*}[!h]
\centering
\caption{The black-box attack success rates (\%) on AgeDB-30 dataset under $\epsilon=16$. The values with a gray background were obtained using the proposed methods. The top-1 best results are highlighted in \textbf{bold}.}
\label{tab1_age}
\setlength{\tabcolsep}{2.8mm}{
\resizebox{1.0\linewidth}{!}{
\begin{tabular}{c|ccccccccc} 

 \hline
 \multirow{3}{*}{Attacks} & TF-NAS-A & GhostNet & SwinT & MFNet & MFNet & MFNet & In-Res-v1 & ResNet50-I-S & Sphere \\
   & @MV & @MV & @MV & @MV & @MV & @Arc & @FN & @Arc & @SF \\
   & @MS & @MS & @MS & @MS & @CA & @CA & @VG & @MS & @CA \\
\hline
FGSM       & 45.90 & 48.06 & 30.93 & 53.43 & 59.10 & 49.74 & 34.86 & 45.85 & 37.28 \\
\rowcolor{darkgray}
FGSM-LFAP  & 55.86 & 53.80 & 42.96 & 57.19 & 56.63 & 54.41 & 48.00 & 52.31 & 45.58 \\
\rowcolor{darkgray}
FGSM-LMFAP & 79.86 & 78.53 & 68.16 & 81.76 & 81.03 & 79.54 & 64.43 & 80.68 & 67.71 \\
\hline
MI       & 31.89 & 34.59 & 18.76 & 38.33 & 43.66 & 34.78 & 21.99 & 32.01 & 25.11 \\
\rowcolor{darkgray}
MI-LFAP  & 61.19 & 58.43 & 47.26 & 62.33 & 63.63 & 59.61 & 51.83 & 56.51 & 51.85 \\
\rowcolor{darkgray}
MI-LMFAP & 88.13 & 86.00 & 78.23 & 88.76 & 86.06 & 88.68 & 71.63 & 88.55 & 79.11 \\
\hline
TI       & 22.36 & 24.43 & 12.26 & 26.33 & 39.00 & 21.78 & 24.50 & 21.81 & 19.85 \\
\rowcolor{darkgray}
TI-LFAP  & 34.06 & 35.83 & 21.09 & 36.23 & 45.99 & 33.71 & 46.93 & 30.48 & 33.48 \\
\rowcolor{darkgray}
TI-LMFAP & 66.43 & 68.96 & 47.86 & 69.33 & 79.13 & 66.58 & 68.03 & 67.68 & 63.31 \\
\hline
SI       & 40.10 & 42.83 & 24.43 & 47.33 & 51.36 & 44.88 & 31.93 & 39.15 & 41.25 \\
\rowcolor{darkgray}
SI-LFAP  & 61.19 & 60.30 & 46.00 & 62.63 & 66.23 & 62.38 & 56.16 & 59.05 & 55.08 \\
\rowcolor{darkgray}
SI-LMFAP & \textbf{92.13} & \textbf{90.06} & \textbf{84.20} & \textbf{91.46} & 87.60 & \textbf{91.41} & \textbf{76.70} & \textbf{92.21} & \textbf{82.65} \\
\hline
DI-MI       & 38.96 & 42.99 & 24.23 & 46.26 & 52.10 & 43.85 & 28.89 & 38.31 & 32.85 \\
\rowcolor{darkgray}
DI-MI-LFAP  & 62.70 & 59.86 & 49.60 & 63.30 & 64.53 & 62.21 & 54.73 & 59.15 & 54.04 \\
\rowcolor{darkgray}
DI-MI-LMFAP & 89.83 & 87.50 & 80.83 & 89.43 & 86.89 & 89.68 & 74.06 & 89.55 & 80.54 \\
\hline
ADMIX      & 32.53 & 36.53 & 19.36 & 40.93 & 49.33 & 39.14 & 21.60 & 33.25 & 28.21 \\
\rowcolor{darkgray}
ADMIX-LFAP & 59.06 & 25.26 & 44.63 & 60.96 & 65.03 & 61.16 & 50.30 & 55.71 & 52.08 \\
\rowcolor{darkgray}
ADMIX-LMFAP& 90.06 & 87.76 & 80.19 & 89.46 & 87.50 & 90.08 & 72.83 & 89.91 & 80.78 \\
\hline
ADMIX-DI-MI      & 48.46 & 50.93 & 32.36 & 55.09 & 62.56 & 54.05 & 35.40 & 50.38 & 40.61 \\
\rowcolor{darkgray}
ADMIX-DI-MI-LFAP & 64.36 & 63.33 & 51.66 & 65.73 & 68.19 & 65.85 & 56.30 & 61.25 & 56.45 \\
\rowcolor{darkgray}
ADMIX-DI-MI-LMFAP& 91.33 & 89.13 & 83.36 & 90.66 & \textbf{87.83} & 91.11 & 75.50 & 90.94 & 82.15 \\
\hline
\end{tabular}}}
\end{table*}

\begin{table*}[!h]
\centering
\caption{The black-box attack success rates (\%) on CFP-FP dataset under $\epsilon=16$. The values with a gray background were obtained using the proposed methods. The top-1 best results are highlighted in \textbf{bold}.}
\label{tab1_cfp}
\setlength{\tabcolsep}{2.8mm}{
\resizebox{1.0\linewidth}{!}{
\begin{tabular}{c|ccccccccc} 

 \hline
 \multirow{3}{*}{Attacks} & TF-NAS-A & GhostNet & SwinT & MFNet & MFNet & MFNet & In-Res-v1 & ResNet50-I-S & Sphere \\
  & @MV & @MV & @MV & @MV & @MV & @Arc & @FN & @Arc & @SF \\
  & @MS & @MS & @MS & @MS & @CA & @CA & @VG & @MS & @CA \\
\hline
FGSM       & 29.27 & 31.18 & 18.85 & 26.88 & 42.97 & 28.42 & 24.91 & 24.58 & 8.89 \\
\rowcolor{darkgray}
FGSM-LFAP  & 35.38 & 36.58 & 24.19 & 30.57 & 36.77 & 31.71 & 32.48 & 30.01 & 22.15 \\
\rowcolor{darkgray}
FGSM-LMFAP & 51.70 & 49.92 & 38.31 & 45.39 & 59.91 & 45.51 & 42.94 & 50.69 & 27.32 \\
\hline
MI       & 22.75 & 25.49 & 12.57 & 21.45 & 30.20 & 21.28 & 15.11 & 18.21 & 13.32 \\
\rowcolor{darkgray}
MI-LFAP  & 40.41 & 41.15 & 26.94 & 36.17 & 43.91 & 35.82 & 34.85 & 36.84 & 26.52 \\
\rowcolor{darkgray}
MI-LMFAP & 60.29 & 57.72 & 47.34 & 54.54 & 70.97 & 55.54 & 49.08 & 63.15 & 36.15 \\
\hline
TI       & 18.78 & 23.01 & 10.71 & 16.62 & 27.65 & 14.91 & 16.02 & 14.04 & 14.41 \\
\rowcolor{darkgray}
TI-LFAP  & 24.50 & 28.44 & 13.91 & 20.08 & 27.62 & 20.94 & 28.82 & 18.92 & 17.92 \\
\rowcolor{darkgray}
TI-LMFAP & 41.84 & 44.84 & 27.68 & 36.25 & 57.71 & 34.51 & 44.40 & 40.90 & 26.44 \\
\hline
SI       & 24.50 & 28.27 & 15.51 & 23.22 & 32.39 & 25.22 & 19.11 & 20.75 & 13.18 \\
\rowcolor{darkgray}
SI-LFAP  & 42.72 & 43.72 & 28.91 & 39.88 & 46.54 & 39.22 & 38.39 & 39.90 & 29.81 \\
\rowcolor{darkgray}
SI-LMFAP & \textbf{67.95} & \textbf{64.90} & \textbf{56.45} & \textbf{62.22} & 77.80 & \textbf{63.31} & 57.42 & \textbf{72.55} & \textbf{42.44} \\
\hline
DI-MI       & 26.90 & 30.30 & 15.88 & 24.11 & 40.25 & 24.34 & 20.14 & 21.44 & 15.58 \\
\rowcolor{darkgray}
DI-MI-LFAP  & 40.64 & 42.44 & 28.82 & 37.28 & 46.51 & 37.34 & 37.71 & 39.72 & 28.35 \\
\rowcolor{darkgray}
DI-MI-LMFAP & 62.95 & 59.09 & 50.20 & 56.68 & 73.99 & 57.94 & 51.85 & 66.01 & 38.12 \\
\hline
ADMIX      & 24.58 & 27.27 & 13.28 & 22.85 & 35.80 & 23.14 & 16.28 & 20.55 & 16.30 \\
\rowcolor{darkgray}
ADMIX-LFAP & 41.18 & 41.69 & 27.48 & 36.34 & 47.00 & 36.88 & 35.85 & 37.15 & 28.75 \\
\rowcolor{darkgray}
ADMIX-LMFAP& 64.75 & 60.84 & 51.54 & 58.37 & 76.05 & 59.77 & 52.34 & 67.98 & 40.72 \\
\hline
ADMIX-DI-MI      & 32.18 & 35.41 & 20.74 & 30.85 & 49.28 & 30.40 & 24.60 & 30.81 & 20.12 \\
\rowcolor{darkgray}
ADMIX-DI-MI-LFAP & 43.58 & 44.67 & 31.42 & 39.68 & 49.97 & 39.14 & 40.77 & 41.92 & 30.38 \\
\rowcolor{darkgray}
ADMIX-DI-MI-LMFAP& 66.01 & 62.49 & 56.11 & 59.65 & \textbf{78.60} & 61.71 & \textbf{57.48} & 71.75 & 41.70 \\
\hline
\end{tabular}}}
\end{table*}

\begin{table*}[hb]
\centering
\caption{The black-box attack success rates (\%) on LFW dataset under $\epsilon=8$. The values with a gray background were obtained using the proposed methods. The top-1 best results are highlighted in \textbf{bold}.}
\label{tab1_lfw_8}
\resizebox{1.0\linewidth}{!}{
\setlength{\tabcolsep}{2.8mm}{
\begin{tabular}{c|cccccccccc} 

\hline
 \multirow{3}{*}{Attacks} & TF-NAS-A & GhostNet & SwinT & MFNet & MFNet & MFNet & In-Res-v1 & ResNet50-I-S & Sphere \\
  & @MV & @MV & @MV & @MV & @MV & @Arc & @FN & @Arc & @SF \\
  & @MS & @MS & @MS & @MS & @CA & @CA & @VG & @MS & @CA \\
\hline
FGSM       & 8.95 & 10.03 & 3.35 & 10.23 & 33.06 & 18.55 & 12.16 & 9.38 & 22.14 \\
\rowcolor{darkgray}
FGSM-LFAP  & 7.35 & 8.03 & 3.31 & 7.33 & 19.00 & 13.28 & 17.00 & 6.64 & 17.68 \\
\rowcolor{darkgray}
FGSM-LMFAP & 23.65 & 24.16 & 12.21 & 22.39 & 51.26 & 36.51 & 31.93 & 26.54 & 39.05 \\
\hline
MI       & 4.58 & 5.30 & 1.58 & 5.26 & 19.86 & 9.58 & 6.36 & 4.34 & 12.91 \\
\rowcolor{darkgray}
MI-LFAP  & 7.84 & 8.30 & 3.78 & 7.90 & 21.13 & 15.18 & 17.69 & 7.31 & 18.28 \\
\rowcolor{darkgray}
MI-LMFAP & 31.25 & 31.50 & 16.91 & 28.93 & 61.86 & 44.81 & 37.83 & 33.94 & 46.45 \\
\hline
TI       & 2.31 & 2.86 & 0.95 & 2.56 & 14.26 & 4.15 & 7.36 & 2.54 & 8.71 \\
\rowcolor{darkgray}
TI-LFAP  & 2.34 & 3.03 & 0.88 & 2.30 & 8.83 & 3.91 & 11.43 & 2.01 & 8.38 \\
\rowcolor{darkgray}
TI-LMFAP & 7.11 & 9.46 & 3.15 & 8.23 & 29.56 & 13.25 & 24.46 & 9.18 & 20.45 \\
\hline
SI       & 4.58 & 5.46 & 1.61 & 5.90 & 19.16 & 9.35 & 8.55 & 4.55 & 17.08 \\
\rowcolor{darkgray}
SI-LFAP  & 6.55 & 7.46 & 2.81 & 6.86 & 19.60 & 12.98 & 18.40 & 6.55 & 18.01 \\
\rowcolor{darkgray}
SI-LMFAP & 34.91 & 35.56 & 17.51 & 33.26 & 66.80 & 50.68 & \textbf{44.96} & 39.01 & 52.78 \\
\hline
DI-MI    & 6.78 & 7.23 & 1.98 & 7.46 & 27.50 & 14.54 & 8.26 & 6.85 & 17.18 \\
\rowcolor{darkgray}
DI-MI-LFAP  & 7.68 & 8.09 & 3.58 & 8.06 & 21.40 & 14.65 & 19.13 & 7.51 & 19.38 \\
\rowcolor{darkgray}
DI-MI-LMFAP & 32.34 & 32.16 & 17.08 & 30.16 & 63.56 & 46.25 & 39.33 & 34.41 & 47.78 \\
\hline
ADMIX      & 6.28 & 6.63 & 1.91 & 6.80 & 26.73 & 13.71 & 7.03 & 6.21 & 15.95 \\
\rowcolor{darkgray}
ADMIX-LFAP & 7.55 & 7.73 & 3.25 & 7.53 & 21.30 & 14.18 & 17.23 & 7.04 & 18.38 \\
\rowcolor{darkgray}
ADMIX-LMFAP& 35.11 & 34.99 & 18.68 & 33.59 & 67.66 & 51.58 & 39.96 & 37.31 & 51.51 \\
\hline
ADMIX-DI-MI      & 11.88 & 13.29 & 4.01 & 14.49 & 41.26 & 23.78 & 12.06 & 13.35 & 25.38 \\
\rowcolor{darkgray}
ADMIX-DI-MI-LFAP & 9.05 & 9.43 & 3.94 & 8.66 & 23.83 & 16.41 & 20.99 & 8.65 & 21.51 \\
\rowcolor{darkgray}
ADMIX-DI-MI-LMFAP& \textbf{36.38} & \textbf{36.60} & \textbf{19.78} & \textbf{35.39} & \textbf{68.56} & \textbf{52.28} & 44.59 & \textbf{39.58} & \textbf{53.01} \\
\hline
\end{tabular}}}
\end{table*}

\begin{table*}[!h]
\centering
\caption{The black-box attack success rates (\%) on AgeDB-30 dataset under $\epsilon=8$. The values with a gray background were obtained using the proposed methods. The top-1 best results are highlighted in \textbf{bold}.}
\label{tab1_age_8}
\resizebox{1.0\linewidth}{!}{
\setlength{\tabcolsep}{2.8mm}{
\begin{tabular}{c|ccccccccc} 

 \hline
 \multirow{3}{*}{Attacks} &  TF-NAS-A & GhostNet & SwinT & MFNet & MFNet & MFNet & In-Res-v1 & ResNet50-I-S & Sphere \\
  &  @MV & @MV & @MV & @MV & @MV & @Arc & @FN & @Arc & @SF \\
  &  @MS & @MS & @MS & @MS & @CA & @CA & @VG & @MS & @CA \\
\hline
FGSM       & 23.86 & 27.13 & 13.19 & 30.20 & 37.86 & 27.98 & 19.13 & 24.71 & 22.21 \\
\rowcolor{darkgray}
FGSM-LFAP  & 21.63 & 22.23 & 13.46 & 24.90 & 31.66 & 24.15 & 28.33 & 18.71 & 23.78 \\
\rowcolor{darkgray}
FGSM-LMFAP & 51.46 & 52.66 & 32.03 & 56.40 & 64.83 & 53.88 & 45.66 & 53.58 & 54.54 \\
\hline
MI       & 16.63 & 19.60 & 8.50 & 21.90 & 28.96 & 19.71 & 12.76 & 17.01 & 16.84 \\
\rowcolor{darkgray}
MI-LFAP  & 22.00 & 23.26 & 12.86 & 25.36 & 32.50 & 25.08 & 27.73 & 19.61 & 24.48 \\
\rowcolor{darkgray}
MI-LMFAP & 56.73 & 58.93 & 36.93 & 62.10 & 71.56 & 60.25 & 49.60 & 59.51 & 59.01 \\
\hline
TI       & 11.66 & 13.63 & 5.63 & 14.70 & 24.83 & 11.55 & 15.76 & 10.74 & 14.35 \\
\rowcolor{darkgray}
TI-LFAP  & 11.20 & 12.29 & 5.79 & 14.26 & 20.10 & 12.31 & 23.06 & 6.61 & 14.25 \\
\rowcolor{darkgray}
TI-LMFAP & 29.96 & 35.36 & 15.93 & 35.46 & 50.36 & 31.45 & 42.26 & 31.38 & 38.95 \\
\hline
SI       & 21.06 & 24.16 & 9.86 & 26.43 & 34.36 & 23.78 & 17.79 & 20.38 & 24.74 \\
\rowcolor{darkgray}
SI-LFAP  & 21.59 & 23.13 & 11.89 & 25.63 & 32.29 & 23.28 & 29.96 & 18.95 & 25.81 \\
\rowcolor{darkgray}
SI-LMFAP & 59.00 & 62.06 & 37.00 & 63.80 & 73.66 & 64.61 & 54.76 & 62.08 & 63.05 \\
\hline
DI-MI       & 20.26 & 21.13 & 10.89 & 27.43 & 33.79 & 23.68 & 16.76 & 21.68 & 23.55 \\
\rowcolor{darkgray}
DI-MI-LFAP  & 22.66 & 23.66 & 13.29 & 26.19 & 33.66 & 24.78 & 30.86 & 20.35 & 27.35 \\
\rowcolor{darkgray}
DI-MI-LMFAP & 58.86 & 59.96 & 38.19 & 63.46 & 73.00 & 61.41 & 53.10 & 62.14 & 61.11 \\
\hline
ADMIX      & 21.96 & 24.63 & 10.30 & 27.23 & 38.86 & 24.31 & 15.80 & 22.08 & 24.28 \\
\rowcolor{darkgray}
ADMIX-LFAP & 20.79 & 22.16 & 12.50 & 25.46 & 33.40 & 24.65 & 28.33 & 19.48 & 25.24 \\
\rowcolor{darkgray}
ADMIX-LMFAP& 60.06 & 60.96 & 39.06 & 64.86 & 75.20 & 63.41 & 51.90 & 62.78 & 63.71 \\
\hline
ADMIX-DI-MI      & 31.86 & 35.76 & 16.60 & 37.73 & 50.83 & 35.98 & 23.70 & 31.78 & 33.31 \\
\rowcolor{darkgray}
ADMIX-DI-MI-LFAP & 25.13 & 26.13 & 15.00 & 28.53 & 37.56 & 27.68 & 31.50 & 22.11 & 29.31 \\
\rowcolor{darkgray}
ADMIX-DI-MI-LMFAP& \textbf{61.73} & \textbf{64.56} & \textbf{41.06} & \textbf{67.59} & \textbf{76.96} & \textbf{65.18} & \textbf{55.06} & \textbf{65.28} & \textbf{66.31} \\
\hline
\end{tabular}}}
\end{table*}

\begin{table*}[!h]
\centering
\caption{The black-box attack success rates (\%) on CFP-FP dataset under $\epsilon=8$. The values with a gray background were obtained using the proposed methods. The top-2 best results are highlighted in \textbf{bold}.}
\label{tab1_cfp_8}
\resizebox{1.0\linewidth}{!}{
\setlength{\tabcolsep}{2.8mm}{
\begin{tabular}{c|ccccccccc} 

 \hline
 \multirow{3}{*}{Attacks} & TF-NAS-A & GhostNet & SwinT & MFNet & MFNet & MFNet & In-Res-v1 & ResNet50-I-S & Sphere \\
  & @MV & @MV & @MV & @MV & @MV & @Arc & @FN & @Arc & @SF \\
  & @MS & @MS & @MS & @MS & @CA & @CA & @VG & @MS & @CA \\
\hline
FGSM       & 18.41 & 22.05 & 9.69 & 18.35 & 23.72 & 17.99 & 12.58  & 14.08 & 14.54 \\
\rowcolor{darkgray}
FGSM-LFAP  & 17.47 & 20.51 & 9.04 & 15.84 & 16.35 & 15.22 & 14.84  & 11.77 & 16.88 \\
\rowcolor{darkgray}
FGSM-LMFAP & 28.84 & 33.19 & 17.30 & 27.50 & 37.07 & 25.88 & 23.15  & 26.80 & 23.22 \\
\hline
MI       & 14.24 & 17.91 & 7.89 & 13.52 & 16.58 & 13.45 & 8.72 & 10.05 & 13.19 \\
\rowcolor{darkgray}
MI-LFAP  & 18.07 & 21.94 & 9.35 & 15.92 & 16.44 & 15.74 & 14.52 & 12.17 & 16.98 \\
\rowcolor{darkgray}
MI-LMFAP & 32.47 & 36.45 & 19.64 & 30.58 & 42.92 & 29.71 & 25.75 & 31.25 & 25.41 \\
\hline
TI       & 13.18 & 17.11 & 6.61 & 11.67 & 14.10 & 9.77 & 9.15 & 7.82 & 12.54 \\
\rowcolor{darkgray}
TI-LFAP  & 13.27 & 16.40 & 6.21 & 11.04 & 10.29 & 10.17 & 10.98 & 7.74 & 12.77 \\
\rowcolor{darkgray}
TI-LMFAP & 19.90 & 26.20 & 10.95 & 17.98 & 25.15 & 17.05 & 19.98 & 16.08 & 17.82 \\
\hline
SI       & 16.18 & 20.27 & 8.65 & 15.02 & 18.34 & 14.71 & 10.28 & 11.44 & 13.07 \\
\rowcolor{darkgray}
SI-LFAP  & 18.32 & 22.68 & 9.22 & 17.42 & 17.12& 16.37 & 15.18 & 12.94 & 17.85 \\
\rowcolor{darkgray}
SI-LMFAP & \textbf{36.98} & \textbf{40.74} & \textbf{22.22} & \textbf{35.20} & \textbf{48.01} & \textbf{33.88} & \textbf{30.55} & \textbf{36.68} & \textbf{27.71} \\
\hline
DI-MI       & 16.72 & 21.19 & 8.58 & 15.30 & 21.98 & 16.59 & 10.92 & 12.40 & 15.65 \\
\rowcolor{darkgray}
DI-MI-LFAP  & 18.29 & 22.25 & 9.15 & 16.72 & 17.90 & 15.74 & 15.98 & 12.59 & 18.08 \\
\rowcolor{darkgray}
DI-MI-LMFAP & 33.64 & 37.57 & 19.61 & 31.44 & 44.64 & 30.17 & 27.58 & 32.85 & 25.97 \\
\hline
ADMIX      & 16.07 & 19.57 & 8.58 & 15.38 & 18.72 & 16.22 & 9.04 & 11.48 & 15.22 \\
\rowcolor{darkgray}
ADMIX-LFAP & 17.64 & 21.02 & 8.52 & 16.55 & 17.04 & 15.65 & 13.52 & 12.11 & 18.42 \\
\rowcolor{darkgray}
ADMIX-LMFAP& 33.18 & 37.05 & 19.44 & 31.61 & 45.67 & 31.94 & 26.30 & 32.85 & 26.68 \\
\hline
ADMIX-DI-MI      & 20.58 & 24.34 & 10.81 & 19.64 & 28.89 & 20.02 & 13.58 & 17.20 & 18.17 \\
\rowcolor{darkgray}
ADMIX-DI-MI-LFAP & 18.64 & 22.82 & 9.75 & 16.90 & 18.90 & 16.71 & 16.27 & 13.82 & 18.45 \\
\rowcolor{darkgray}
ADMIX-DI-MI-LMFAP& 34.78 & 38.77 & 21.07 & 32.84 & 47.44 & 33.05 & 29.47 & 34.68 & 27.48 \\
\hline
\end{tabular}}}
\end{table*}

\subsection{Experimental Results on Resisting JPEG Attacks}

\subsubsection{Experimental Analysis}\label{sec5-analyses}

In order to substantiate the proposition that confining adversarial perturbations to the low-frequency domain results in attenuated efficacy, we undertook a comprehensive experimental investigation comparing unconstrained adversarial perturbations (Unconstrained) with their low-frequency counterparts delineated in \cite{SharmaDB19} (DCT-Low). To synthesize these adversarial perturbations, we harnessed the capabilities of ResNet18, a model pre-trained on the ImageNet dataset. Following the setup of \cite{SharmaDB19}, we (1) generated adversarial perturbations for all images encompassed within the validation set by employing the FGSM attack strategy, (2) utilized DCT and IDCT techniques to extract high-frequency constituents $[180, 224]$ from the perturbations, and (3) subsequently evaluated the attack success rate on the ResNet50 model. The ResNet50 model was subjected to testing post-compression via JPEG, utilizing a quality factor of 50.

The outcomes of DCT-Low and Unconstrained perturbations across diverse scenarios are depicted in Figure~\ref{fig2-a}. Upon examination, we discerned that (1) the attack success rate of Unconstrained (w/ JPEG on black-box) is inferior to Unconstrained (w/o JPEG on black-box), corroborating the assertion that JPEG compression mitigates the ramifications of adversarial perturbations. (2) The attack success rate of DCT-Low (w/ JPEG on black-box) surpasses that of Unconstrained (w/ JPEG on black-box), insinuating that low-frequency adversarial perturbations are more resilient to JPEG compression than their original counterparts. Nonetheless, (3) in the absence of JPEG compression, the attack success rate of DCT-Low (w/o JPEG on black-box) is suboptimal in comparison to Unconstrained (w/o JPEG on black-box). (4) In real-world applications, wherein generated adversarial images are subjected to JPEG compression and assessed on black-box models, DCT-Low (w/ JPEG on black-box) exhibits superior performance relative to Unconstrained (w/ JPEG on black-box). However, in interference-free circumstances, DCT-Low (w/o JPEG on white-box) underperforms in comparison to Unconstrained (w/o JPEG on white-box). These findings intimate that confining the adversarial perturbations undermines their effectiveness in certain contexts. A more viable strategy entails reconstructing the source model to leverage low-frequency data and harnessing it to generate low-frequency adversarial perturbations.

\subsubsection{Comparison with Resisting JPEG Attacks}

To demonstrate the performance improvement of our methods compared to existing methods that resist JPEG compression, we compare our method with $6$ baseline attacks on MFNet@MV@CA under $\epsilon=16$, and the results of generated adversarial images after the JPEG compression with different quality factors are shown in Table~\ref{tab2}. It should be noted that in order to get fair comparison results, the adversarial attack algorithms we employed in Table~\ref{tab2} to generate adversarial examples are ADMIX-DI-MI. We can observe that: (1) The method tackling target JPEG compression of specific factors underperforms when faced with other factors. \emph{E.g.,} Differentiable-JPEG-75 achieves worse performance without JPEG compression than when the compression quality factor is 75. Moreover, attacks designed specifically for JPEG compression perform worse than algorithms that focus only on improving the attack success rates, suggesting that confining noise to circumvent the impact of JPEG compression is not a sound solution. (2) LFAP achieves convincing performance, and LMFAP outperforms all baselines in all situations. (3) The stronger performance of LMFAP over LFAP indicates that our incorporation strategy is effective for different compression qualities.

\begin{table}[t]
\centering
\caption{The evaluation of \texttt{Face++} API with the different JPEG compression on CASIA-WebFace dataset under $\epsilon=16$.  The top-1 best results are highlighted in \textbf{bold}.}
\label{tab3}
\setlength{\tabcolsep}{0.4mm}{
\begin{tabular}{c|ccc} 

 \hline
  Attacks & W/O & Quality=75 & Quality=50 \\
  \hline
  Differentiable-JPEG-75 & 34.89 & 36.33 & 36.13 \\
  Differentiable-JPEG-50 & 37.55 & 36.74 & 40.21 \\
  ComReAdv-75            & 23.87 & 24.90 & 25.72 \\
  ComReAdv-50            & 25.30 & 26.53 & 26.74 \\
  TAP                    & 26.12 & 25.31 & 30.41 \\
  DCT-Low                & 36.12 & 34.29 & 39.19 \\
  LowKey                 & 51.63 & 51.43 & 52.04 \\
  APF                    & 65.30 & 65.11 & 67.96 \\
  \hline
  FGSM                   & 46.53 & 47.35 & 50.82 \\
  \rowcolor{darkgray}
  FGSM-LMFAP             & 51.02 & 49.80 & 52.04 \\
  \hline
  MI                   & 23.26 & 20.41 & 24.29 \\
  \rowcolor{darkgray}
  MI-LMFAP             & 61.22 & 61.64 & 63.27 \\
  \hline
  TI                   & 15.10 & 14.29 & 14.7 \\
  \rowcolor{darkgray}
  TI-LMFAP             & 39.79 & 39.80 & 43.27 \\
  \hline
  SI                   & 50.40 & 48.78 & 48.37 \\
  \rowcolor{darkgray}
  SI-LMFAP             & \textbf{72.65} & \textbf{74.49} & \textbf{74.29} \\
  \hline
  DI-MI                   & 28.57 & 27.96 & 29.39 \\
  \rowcolor{darkgray}
  DI-MI-LMFAP             & 65.51 & 64.49 & 66.53 \\
  \hline
  ADMIX                   & 23.06 & 22.45 & 23.06 \\
  \rowcolor{darkgray}
  ADMIX-LMFAP             & 65.91 & 65.51 & 67.15 \\
  \hline
  ADMIX-DI-MI                   & 34.69 & 35.72 & 35.92 \\
  \rowcolor{darkgray}
  ADMIX-DI-MI-LMFAP             & \textbf{70.20} & \textbf{67.96} & \textbf{69.60} \\

\hline
\end{tabular}}
\end{table}

\begin{figure}[t]
  \centering
  \includegraphics[width=\linewidth]{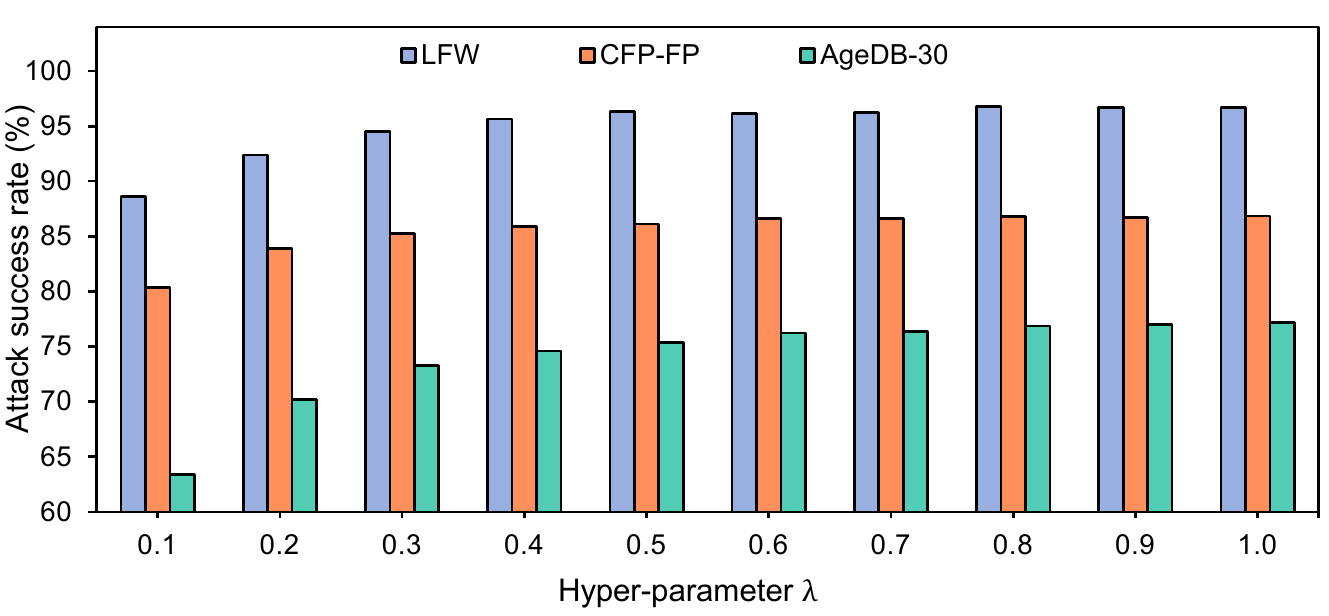}
  \caption{Effects of $\lambda$ under $\epsilon=16$. $\lambda = 0.6$ is used as default.}\label{fig5}
\end{figure}

\begin{figure}[t]
  \centering
  \includegraphics[width=\linewidth]{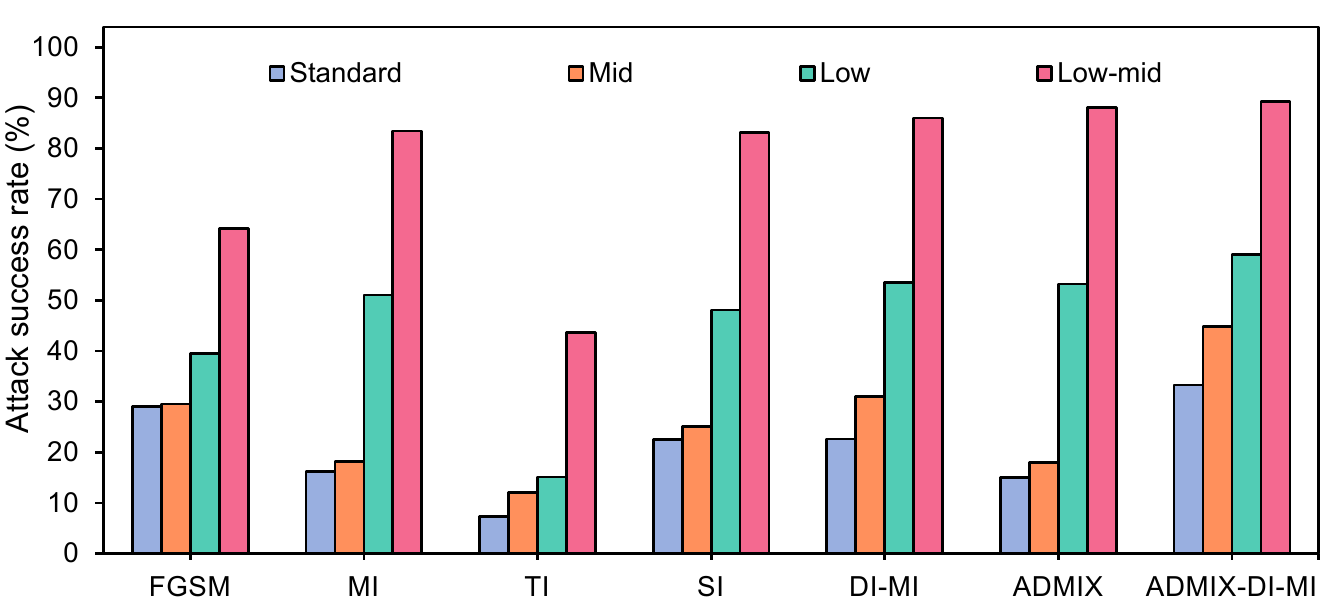}
  \caption{Effects of different frequency components. TF-NAS-A@MV@MS is used as the black-box model. LFW is used as the dataset.}\label{fig_mid}
\end{figure}

\subsection{Experimental Results on Black-box Attacks}\label{sec5-3}

In order to evince the compatibility of our methods with existing black-box attacks and to augment their efficacy, we conducted an appraisal of seven baseline attacks, as well as a study of attacks on nine FRS under the suggested threshold, \emph{i.e.}, $\epsilon=16$. The results of the generated adversarial images were obtained from the LFW, AgeDB-30, and CFP-FP datasets, and are presented in Table~\ref{tab1}, Table~\ref{tab1_age} and Table~\ref{tab1_cfp}, respectively. We also show the experimental results under $\epsilon=8$ in Table~\ref{tab1_lfw_8}, Table~\ref{tab1_age_8} and Table~\ref{tab1_cfp_8}, respectively. We have the following main findings: (1) Our method, LMFAP incorporated with the state-of-the-art black-box attacks, \emph{i.e.}, ADMIX-DI-MI or SI, achieves the best performance than all existing black-box attacks. \emph{E.g.}, in Table~\ref{tab1}, ADMIX-DI-MI-LMFAP yields the attack success rates that are almost always above $80.0\%$, even a margin of $39.74\%$ than the state-of-the-art ADMIX-DI-MI on average. In other tables, \emph{e.g.}, Table~\ref{tab1_age}, the LMFAP based on SI and ADMIX-DI-MI together are the best approaches. Our solution achieves the best performance in all tables. This demonstrates the performance of our methods in resisting JPEG compression and facing unknown FRS. (2) Our methods (both LFAP and LMFAP) are compatible with various black-box attacks. Almost all black-box attacks can significantly improve performance when incorporated with our methods (both LFAP and LMFAP). The stronger the black-box attack, the better performance the incorporated attack achieves. This motivates us to use more effective black-box attacks when deploying the incorporated framework in practical applications. (3) LMFAP basically outperforms LFAP, which proves that our incorporation strategy is effective. (4) By comparing MFNet@MV@MS, MFNet@MV@CA and MFNet@Arc@CA, rather than the different backbones, we find that the other different black-box settings in the FRS also have an effect on the black-box attacks, \emph{i.e.,} the training set and the head. This is an undiscussed finding for researchers studying black-box attacks in FRS.

Figure~\ref{fig2-b} qualitatively visualizes the generated adversarial images. Figure~\ref{fig_ssim} quantitatively shows the corresponding SSIM values for the whole dataset. From the qualitative results, all the adversarial images show very limited interference to the human eye, and neither LFAP nor LMFAP significantly increase the interferability, compared to the raw baseline attacks. This is also confirmed by the quantitative results, where the images generated by LFAP and LMFAP are very close to the original attacks in terms of SSIM values and both are well above the 0.8 criterion for measuring high-quality images~\cite{zhang2020adversarial}.

\subsection{Experimental Results on Commercial Black-box API}\label{sec5-4}

\begin{table*}[t]
\centering
\caption{Statistical attack success rates of the five trial runs. MFNet@MV@MS is used as the black-box model. LFW is used as the dataset.}
\label{tab_statistical}
\setlength{\tabcolsep}{0.8mm}{
\begin{tabular}{c|cc} 

 \hline
 Attacks & Mean & Variance  \\
  \hline
  FGSM       & 31.32 & 0.3491 \\
  \rowcolor{darkgray}
  FGSM-LFAP  & 34.32 & 0.5165 \\
  \rowcolor{darkgray}
  FGSM-LMFAP & 60.98 & 0.8702 \\
  \hline
  MI       & 17.38 & 0.5328 \\
  \rowcolor{darkgray}
  MI-LFAP  & 44.24 & 0.6068 \\
  \rowcolor{darkgray}
  MI-LMFAP & 79.28 & 0.3259 \\
\hline
  TI       &  7.61 &  0.3003 \\
  \rowcolor{darkgray}
  TI-LFAP  & 13.65 & 0.4671 \\
  \rowcolor{darkgray}
  TI-LMFAP & 37.98 & 0.3904 \\
  \hline
  SI       & 28.20 & 0.9279 \\
  \rowcolor{darkgray}
  SI-LFAP  & 41.97 & 1.0637 \\
  \rowcolor{darkgray}
  SI-LMFAP & 77.91 & 1.3798 \\
  \hline
  DI-MI       & 22.78 & 0.7335 \\
  \rowcolor{darkgray}
  DI-MI-LFAP  & 45.15 & 1.1690 \\
  \rowcolor{darkgray}
  DI-MI-LMFAP & 81.91 & 0.9185 \\
  \hline
  ADMIX      & 18.23 & 0.6199 \\
  \rowcolor{darkgray}
  ADMIX-LFAP & 45.40 & 0.8814 \\
  \rowcolor{darkgray}
  ADMIX-LMFAP& 84.33 & 0.7369 \\
\hline
  ADMIX-DI-MI      & 34.42 & 0.8101 \\
  \rowcolor{darkgray}
  ADMIX-DI-MI-LFAP & 49.95 & 1.0184 \\
  \rowcolor{darkgray}
  ADMIX-DI-MI-LMFAP& 86.19 & 0.8890 \\
\hline
\end{tabular}}
\end{table*}

Our method was assessed against the real-world commercial black-box API: \texttt{Face++}. This API is widely employed in industrial applications and its exact specifications are not publicly available. We randomly selected 10 subjects from the CASIA-WebFace dataset, each with 50 face images. Among these images, one was randomly chosen as the gallery image (images with known identities) while the remaining 49 images were used as probe images (images intended to be identified). It is significant to note that \texttt{Face++} does not divulge its algorithms, including models, datasets, or even potential image pre-processing techniques. Hence, the evaluation results of our method on this API carry critical practical implications. We compared our methods with all baseline attacks on the \texttt{Face++} API, as presented in Table~\ref{tab3}. We noted that even in this situation, our methods (SI-LMFAP and ADMIX-DI-MI-LMFAP) outperformed all other baseline methods, consistently with Section~\ref{sec5-3}. Our experimental findings on our designed black-box settings and commercial black-box APIs demonstrate the generality of our solution, regardless of the scenario.

\subsection{Ablation Study}\label{sec5-5}

\subsubsection{Influence of Hyper-parameter}
We have shown the superior performance of LMFAP when $\lambda = 0.6$. The influence of hyper-parameter $\lambda$ has not yet been explored. Taking MFNet@MV@CA as the target model, we evaluated ADMIX-DI-MI-LMFAP under different values of $\lambda$ as shown in Figure~\ref{fig5}. When $\lambda>0.3$, the performance of the attack is almost optimal on all three datasets, suggesting that this hyper-parameter needs no careful tuning. Moreover, even the worst value ($\lambda=0.1$) performs better than baselines.

\subsubsection{Influence of Mid-frequency Components}
In this work, we have synthesized the low-frequency components with the mid-frequency components synergistically to attain favorable outcomes. However, it remains unexplored whether this combination has served as a valuable complement to both. To address this, we conducted a comparative analysis of the performance of noise at standard frequency (derived from the standard trained FRS), noise at mid frequency (derived from the subprime robust FRS $f_2$), noise at low frequency (derived from the prime robust FRS $f_1$), and noise at low-mid frequency (derived from the $f_1$ and $f_2$), as depicted in Figure~\ref{fig_mid}. It is evident that the low-frequency and mid-frequency noise not only outperform the standard noise but also exhibit a heightened efficacy when combined. This substantiates the effectiveness of our proposed strategy.

\subsection{Stability Analysis}\label{sec5-6}

Given that our strategy can be incorporated with certain volatile methods (e.g., DI-MI with integrated input random transformations), it is imperative to assess and explore its stability. We conducted the adversarial image generation procedure five times, and the resultant means and variances are presented in the Table~\ref{tab_statistical}. It is evident that our strategy surpasses the baseline method in terms of mean, with no notable fluctuations observed in the variance. This substantiates the stability of our proposed strategy.

\section{Conclusion}

In this paper, we investigate the efficacy of adversarial perturbations in mitigating the impact of JPEG compression on machine learning models. Initially, we conduct a comprehensive analysis of existing low-frequency perturbation generation methodologies, identifying their limitations, and subsequently develop a more resilient training framework for the construction of the low frequency adversarial perturbation (LFAP). Subsequently, we refine our approach by conducting an in-depth investigation into the entire frequency domain spectrum, subsequently devising the low-mid frequency adversarial perturbation (LMFAP), an innovative adversarial perturbation that synergistically integrates low-frequency and select mid-frequency components. Lastly, we perform rigorous experimentation under various conditions, conclusively demonstrating that our proposed techniques surpass both contemporary methods designed to withstand JPEG compression and extant black-box methodologies.

\section{Acknowledgements}
This work is supported by the Fundamental Research Funds for the Central Universities (No. 2023JBZY033).

\section{Declaration of generative AI and AI-assisted technologies in the writing process}

During the preparation of this work the authors used GPT-4 in order to improve language and readability. After using this tool/service, the authors reviewed and edited the content as needed and take full responsibility for the content of the publication.









\section{}\label{}

\printcredits

\bibliographystyle{cas-model2-names}

\bibliography{cas-refs}



\end{document}